\documentclass[conference]{IEEEtran}
\usepackage{color}
\usepackage{colortbl}

\newcommand\cut[1]{}

\newcommand{\para}[1]{{\vspace{4pt} \bf \noindent #1 \hspace{10pt}}}

\newcommand{\sysname}{{\tt ASF}}
\definecolor{Gray}{gray}{0.9}
\usepackage[dvipsnames]{xcolor}
\usepackage{listings}
\usepackage{makecell}
\usepackage{footnote}
\usepackage{multirow}

\usepackage{tablefootnote}
\newtheorem{definition}{Definition}

\ifCLASSINFOpdf
   \usepackage[pdftex]{graphicx}
\else
   \usepackage[dvips]{graphicx}
\fi
\usepackage[ruled, lined, linesnumbered, noend]{algorithm2e}

\author{\IEEEauthorblockN{Zhisheng Hu}
\IEEEauthorblockA{Baidu Security}
\and
\IEEEauthorblockN{Shengjian Guo}
\IEEEauthorblockA{Baidu Security}
\and
\IEEEauthorblockN{Zhenyu Zhong}
\IEEEauthorblockA{Baidu Security}
\and
\IEEEauthorblockN{Kang Li}
\IEEEauthorblockA{Baidu Security }}

\usepackage{url}

\begin{document}
%

\title{Coverage-based Scene Fuzzing for Virtual Autonomous Driving Testing}

\IEEEoverridecommandlockouts
\makeatletter\def\@IEEEpubidpullup{6.5\baselineskip}\makeatother

\maketitle

\begin{abstract}

Simulation-based virtual testing has become an essential step to ensure the safety of autonomous driving systems. 
%
Testers need to handcraft the virtual driving scenes and configure various environmental settings like surrounding traffic, weather conditions, etc.
%
Due to the huge amount of configuration possibilities, the human
efforts are subject to the inefficiency in detecting flaws in industry-class autonomous driving system.
This paper proposes a coverage-driven fuzzing technique to automatically generate diverse
configuration parameters to form new driving scenes. 
Experimental results show that our fuzzing method can significantly
reduce the cost in deriving new risky scenes from the initial setup designed by testers.
We expect automated fuzzing will become a common practice in virtual testing for 
autonomous driving systems. 


\end{abstract}


%

\section{Introduction}
Autonomous driving, or auto-driving for short, represents the latest advances in AI-powered transportation while safety acts as the major concern to the success of auto-driving.
%
To test an autonomous driving system's ability of properly handling complicated traffic scenarios, vendors have been running their autonomous driving vehicles with millions of miles of physical road tests~\cite{Apollo_miles,schwall2020waymo}. 
However, despite the gained confidence from the physical world, road 
testing can hardly trigger many rare and dangerous driving scenarios.
For both cost and efficiency reasons, top companies also complement road testing with 3D simulation-based virtual driving testing~\cite{apollosim,scanlon2021waymo}.




With 3D simulators, developers can build multiple virtual test cases to analyze how well
an autonomous driving system responds to abnormal driving scenarios. 
A virtual test case usually consists of a static driving scene configuration and a set of dynamic configurations. The static configuration often includes map, road, and obstacles while
the dynamic configurations can be the surrounding vehicle behaviors, road situations, and weather conditions, etc. 
A common practice of generating virtual test cases is to setup basic static driving scene and dynamic configurations based on reports of real accident scenes ~\cite{scanlon2021waymo}. 
In practice, however, test designers have to choose the set of dynamic configurations by themselves. Though the 3D simulators provide abundant and configurable choices to tune the close-to-reality simulations, a typical accident report merely contains a static scene description rather than such levels of details. In the simulation, the dynamic environmental information like the speeds and positions of surrounding vehicles, the light intensity, and the weather condition are all left to the test designers' experiences. 


Moreover, we observed that the dynamic configurations for 3D simulation can significantly 
affect the testing outcome. 
Similar findings are also reported in recent works~\cite{fremont-itsc20, 9251068}.
To be specific, we found that behavioral variations of surrounding vehicles or light 
intensity changes could cause unexpected accidents. That is, replaced with a certain 
set of dynamic configurations, a benign virtual driving test case can induce remarkable 
safety challenges to an autonomous driving vehicle. 
%
%
Therefore, simulation-based safety testing needs to address an open question -- how to 
find the effective dynamic configurations that contribute to more challenging test cases? 

Driving test case re-construction with more human efforts is by no means the answer. 
%
{Nowadays, software test automation assemblies essential yet repetitive testing tasks 
in automatic ways and performs large-scale test generation that are difficult to construct 
manually.
Similar to software testing practice, virtual driving testing also desires
automated and practical techniques.
}
Toward this end, we propose \sysname, a coverage-driven {\textbf{A}}utonomous {\textbf{S}}afety {\textbf{F}}uzzing technique.
At the high level, \sysname~adopts the idea of software fuzzing testing for the 
test case generation in virtual driving testing. 
Specifically, we design multiple strategies to mutate the changeable content of a 
simulation input --- the dynamic configurations in a driving test case. Given the test 
cases including static driving scenes and the mutated data, \sysname~repeatedly 
executes the simulator with these inputs.
Uniquely, we propose \textit{trajectory coverage}, a metric dedicated for assessing the 
performance of a driving test case. Once the autonomous driving vehicle delivers a new 
driving trajectory in the simulation, \sysname~treats the running test case as a valuable 
candidate and queues it for subsequent mutation. Also, if the autonomous driving vehicle 
encounters a collision, \sysname~treats it as a potential problem and records the running 
test case for human investigation.
Based on the feedback mechanism upon the metric, \sysname~evaluates the newly 
mutated test cases in simulation and keeps on mutating chosen candidates to 
generate more test cases for continuous evaluation, thus forming a 
evolving loop to search for effective dynamic configurations in virtual test cases.
%

We used \sysname~to generate virtual driving test cases upon the open-source version of autonomous driving system Apollo 5.0.0~\cite{Baiduapollo} and the SVL simulator
\cite{LGSVL}. 
Experimental results show that \sysname~generates a desired number of test cases 
within a short search time budget. Running these test case makes the autonomous 
driving vehicle drive significantly diverse trajectories. 
In our experiments, \sysname~also revealed several system flaws that are extremely 
hard to find by random or human crafted test cases. These flaws have all been 
confirmed and fixed in the latest version of Apollo 6.x. 



\section{Practices and Challenges of Virtual Safety Testing for Autonomous Driving}
\label{sec:challenge}

Autonomous driving system senses the environment and plans driving decisions with few or no human intervention. When referring to the safety properties, people usually conduct large-scale road testing 
to fulfill the safety requirements. However, millions of on-road mileage still rarely covers dangerous corner scenarios. In order to make up for the insufficiency of road testing in cost and coverage, people start virtual safety testing for autonomous driving systems through 3D simulators.

%
%
\subsection{Safety testing in virtual simulations}
\para{Virtual safety testing.}
Virtual safety testing supplements physical road testing by creating extensive 
simulation test cases that mimic interested real-world driving scenarios. It can 
repeatedly run and exam the safety properties of the targeted autonomous driving 
system, so as to magnify how well the targeted system responds to the designed 
cases.
To build an effective virtual safety testing technique, the involved simulator should 
serve as the digital mirror of the physical driving environment. The driving simulation 
should produce high-fidelity reflection of what the autonomous driving vehicle encounters 
on the road, especially the abnormal driving scenarios. Next, we will explain how the 
virtual driving simulation works.

\begin{figure}[h]
    \begin{center}
    \includegraphics[width=1\linewidth]{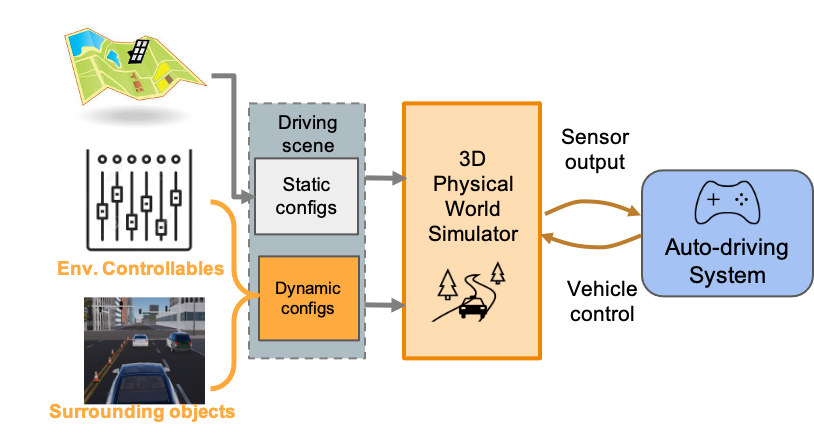}
    \caption{Driving simulation overview}\label{fig:sim}
    \end{center}
\end{figure}

\para{Virtual driving simulation.} 
Fig.~\ref{fig:sim} illustrates the typical flow of a virtual driving simulation.
The simulator utilizes 3D game engines to construct photo-realistic virtual driving scenes that are similar to the ones in the real world. The simulator also provides various sensors, such as LiDAR, camera, IMU, and GPS, to present environment perception in the virtual driving. In addition, the simulator provides a communication bridge that exchanges messages between itself and the targeted autonomous driving system. Then the targeted system is connected to the simulator through  the  communication  bridge  to  control the simulated vehicle.

In this paper, we leverage the  advanced  3D  simulator SVL~\cite{LGSVL} 
and the popular L4 autonomous driving platform Apollo~\cite{Baiduapollo} to demonstrate virtual safety testing.
%
The same behavior logic of Apollo used in the road testing is replicated in the virtual driving simulation: it uses the simulated sensor data provided by SVL to perform the driving task including obstacle prediction, path planning and controlling the vehicle in a given driving scene.

\begin{figure}[h]
    \centering
    \begin{minipage}{0.49\linewidth}
    \centering
    \includegraphics[height=1.4\linewidth]{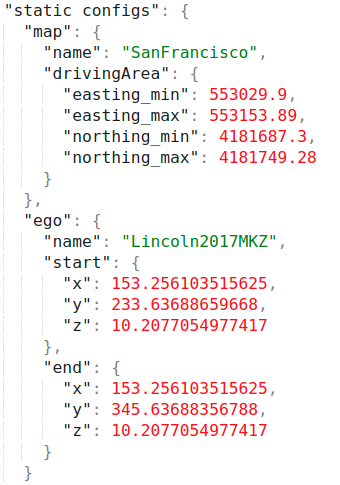}
    {\small (a)~Static configurations}
    \end{minipage}%
    \begin{minipage}{0.49\linewidth}
    \centering
    \includegraphics[width=1.1\linewidth]{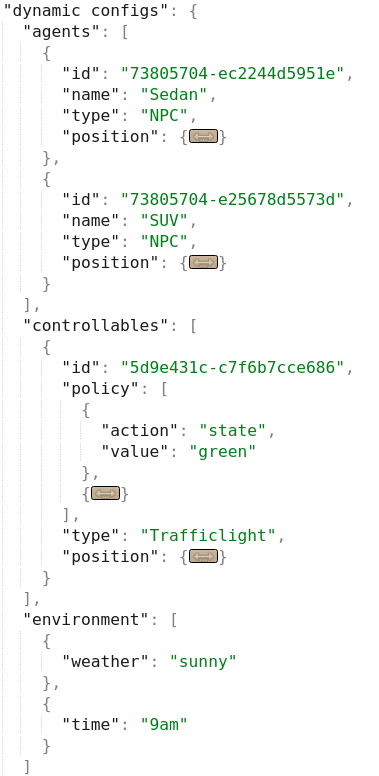}
    {\small (b)~Dynamic configurations}
    \end{minipage}
    \caption{Formalized test case}
    \label{fig:JSON}
\end{figure}

\subsection{Formalized virtual test case}
Fig~\ref{fig:JSON} displays a JSON-format test case of the SVL simulator. This example 
showcases the compact static and dynamic configurations to start a driving environment 
simulation: 

\begin{itemize}
    \item \emph{Static configurations}: a set of configurable elements that sets up 
    the skeleton of a driving scene. These elements include the name of the 
    high-definition map, the driving area on the map, and the start and end points
    of the autonomous driving vehicle.
    
    \item \emph{Dynamic configurations}: the set of elements whose values can be altered in dynamic simulation. These elements include multiple surrounding NPC vehicles, controllable objects of traffic lights, and environment factors like light and weather conditions.
\end{itemize}


The static configurations remain unchanged throughout the entire driving scene simulation. 
For example, the autonomous driving car \texttt{Lincoln2017MKZ} will operate within the 
pre-designed \texttt{drivingArea} on the \texttt{SanFrancisco} map. 

By contrast, altering the dynamic configurations in simulation  can derive more safety-related 
driving situations against the \texttt{ego} car. For example, changing the number, positions, 
and maneuvers of surrounding \texttt{NPC} vehicles challenges the obstacle detection and
path assessment. 
Irregularly flipping the controllable \texttt{Trafficlight} tests the signal recognition 
and error signal resilience.
Increasing the light intensity or deteriorating the weather threatens the perception 
precision.  



\subsection{Challenges of selecting the dynamic configurations}

Existing virtual safety testing methods heavily rely on the scenario database populated 
from accidental reports~\cite{putz2017database,9090897}. These databases often serve as 
the basis for constructing virtual test cases. However, the textual descriptions and 
limited images from the reports are insufficient for creating a vivid 3D simulation, 
and the developers have to artificially fill the missing environmental information \cite{scanlon2021waymo}. 

Then, choosing appropriate dynamic configuration values from the huge search space becomes 
the major challenge in creating effective virtual test cases.  
For example, within the \texttt{drivingArea} of \texttt{SanFrancisco} map, the dynamic 
configurations, such as the maneuvers of surrounding \texttt{NPC} cars, can easily form thousands of possibilities. 
The selection among possible choices can significantly affect the resulting outcome. 
Therefore, finding a practical and automated way to make right selections for virtual 
safety testing is of essential importance.

\section{Apply Fuzzing for Virtual Safety Testing}
We observe that the test case generation for virtual driving testing bears a strong similarity 
with software fuzz testing (fuzzing). 
For example, 
the selection of dynamic configuration values can be analogous to the input mutation in fuzzing; 
the collisions of vehicles or traffic rule violations can be analogous to the program crashes 
found by fuzzing. In what follows, we first introduce the general flow and the core components 
of fuzzing, then describe the challenges of adopting fuzzing for test case generation in virtual 
driving testing. 

\subsection{Fuzz Testing}
Fuzzing~\cite{AFL,Miller95fuzzrevisited,Fuzzing} has achieved significant successes in 
software testing and vulnerability detection, owning to its practicability and scalability 
in code coverage improvement and test case generation. 

\begin{figure}[h]
    \begin{center}
    \includegraphics[width=1\linewidth]{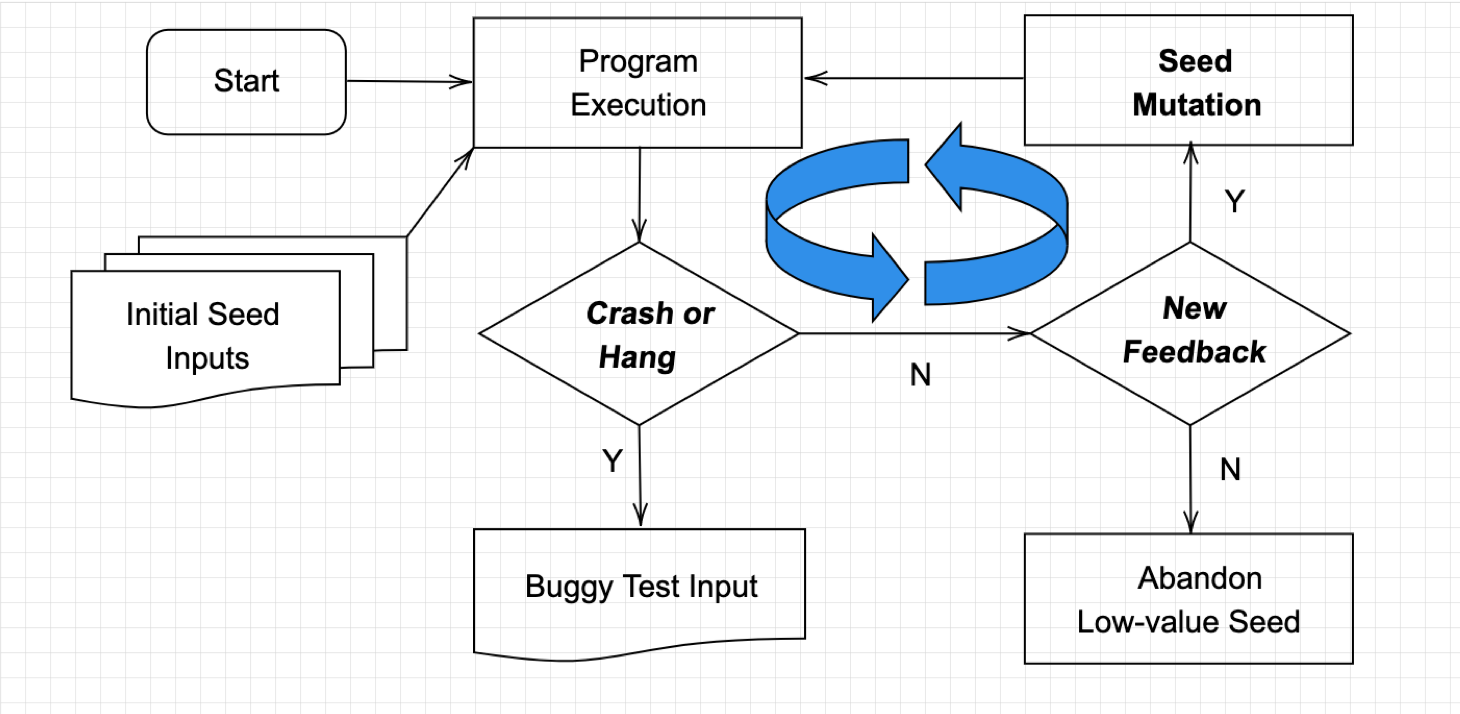}
    \caption{Fuzz testing flow and the core components}\label{fig:flow}
    \end{center}
\end{figure}

Figure~\ref{fig:flow} shows the workflow of a fuzzing engine (i.e., fuzzer). Without loss of 
generality, we separate the fuzzer into three core components including the crash signal 
processing, the feedback mechanism, and the seed mutation, as emphasized by the bold fonts 
in Figure~\ref{fig:flow}. 
Given a set of initial seed input files, the fuzzer repeatedly executes the program under
testing with the initial seeds and later the mutated test inputs originated from these seeds. 
If the fuzzer encounters a program crash or halting in dynamic execution, %
%
it records the bug-triggering input into a persistent file. 
We call this 
action as \emph{crash signal processing}. 
If running an input receives an unique runtime feedback like new code coverage rather than
a crash, the fuzzer would treat the input as a valuable seed and prioritize it as a candidate
for near future mutation. The coverage feedback and the evaluation of this metric form the 
\emph{feedback mechanism}.
Then, based on a queue of prioritized candidate seeds, the fuzzer performs fast seed mutation 
through several strategies like bit and byte flipping, arithmetic increment, value replacement, 
etc. We name this action as \emph{seed mutation}.
On each new input from seed mutation, the fuzzer repeats the program execution, crash signal 
processing, feedback evaluation, and further mutations, thus forming a non-terminating fuzzing 
loop.

\subsection{Challenges of virtual safety fuzzing} \label{sec:fuzzing}

Applying the fuzzing idea to producing diverse dynamic simulation configurations for virtual 
safety testing is imaginative yet challenging. In general, mutating structured parameters is 
close to the intuitions of grammar-based fuzzing~\cite{GodefroidKL08,EberleinN0G20} and 
structure-aware fuzzing~\cite{BlazytkoASASWH19,YouLMPZ019}. The bug symptoms in software 
fuzzing could correspond to safety issues under virtual safety testing. 
However, the primary challenge comes from the feedback mechanism that leads the fuzzing 
progress. Existing fuzzers often utilize low-level code instrumentation to obtain the 
code coverage of the running seed, which guides subsequent seed mutation hence steering 
the fuzzing direction. However, the internal code coverage no longer suits virtual safety
testing since the latter focuses on the behavioral safety. Using the code coverage may 
help find implementation bugs rather than detecting safety risks in simulation.
%

Therefore, virtual safety fuzzing demands a new metric 
that bridges both the autonomous 
driving state and the evolving simulation state. With this metric, we can depict the goal 
and monitor the progress of fuzzing. After running a driving simulation, we evaluate the 
finished driving state against the metric. Then, upon the evaluation feedback, the fuzzing 
conducts pertinent mutations on interested configurations to form new test cases for 
continuous simulations.
\section{A Coverage-based Fuzzing System for Virtual Safety Testing}

\begin{figure*}[h]
    \begin{center}
    \includegraphics[width=1\linewidth]{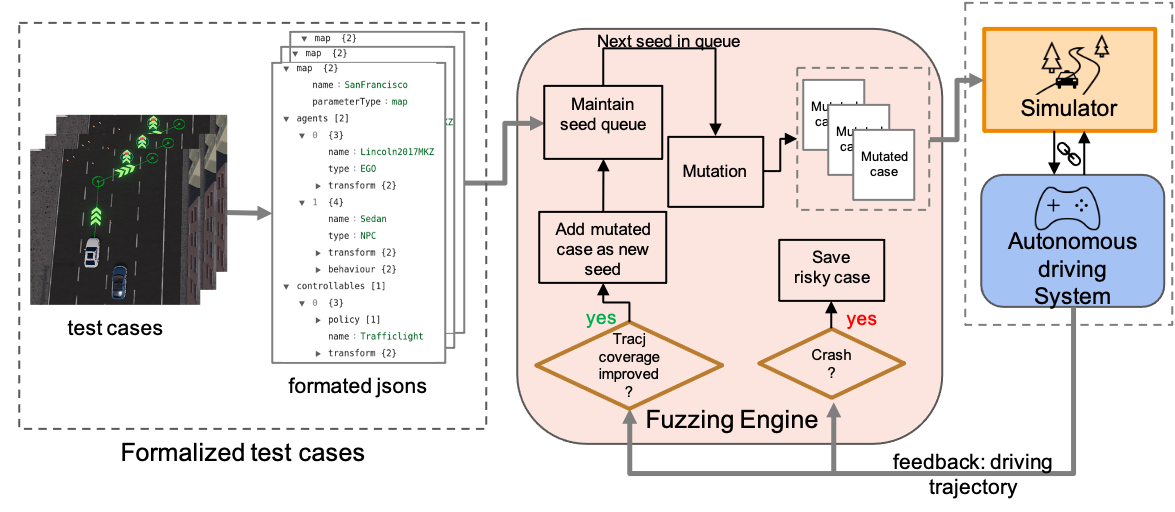}
    \caption{System overview}\label{fig:framework}
    \end{center}
\end{figure*}

We propose \sysname, a coverage-based fuzzing system to efficiently generate risky test cases
for virtual safety testing.

\subsection{The overall flow}
\label{sec:asf_flow}

Fig.~\ref{fig:framework} shows our \sysname~framework, which consists of four components: 
the formalized test cases, the fuzzing engine, the 3D simulator, and the autonomous driving 
system under test. 

%
Briefly speaking, \sysname~works as follows: given a small set of formalized test cases as the 
initial seed inputs, the fuzzing engine repeatedly run the simulation.
%
We call running simulation on a mutated test case as an iteration. At each iteration, the simulator loads a mutated case from fuzzing engine to render a virtual driving scenario. Then, the autonomous 
driving system is bridged into the simulator as the backend controller to drive the specified vehicle in the virtual scenario. 
After an iteration, \sysname~evaluates whether the finished run has increased the trajectory 
coverage --- a new metric defined in Section~\ref{sec:asf-a} dedicated for virtual safety 
fuzzing. If the feedback reports coverage increase, the executed test case would be kept for 
further mutation. Meanwhile, if the driving vehicle encounters an accident or a traffic rule violation, \sysname~records the correlated test case. 
Then, \sysname~resumes execution and proceeds to the next iteration. 
Note that a test case includes static and dynamic configurations in JSON format where we focus 
on the dynamic part (ref. Section~\ref{sec:challenge} ). For brevity, \emph{dynamic configurations} and \emph{test cases} are interchangeable in the rest of the paper.
%


\subsection{The fuzzing algorithm}

Algorithm~\ref{algo:search} formalizes the workflow of \sysname's fuzzing engine.  
The algorithm starts with some randomly generated test cases by placing them into 
the priority queue $\mathcal{Q}$ as the initial seeds.
Then, it enters the first $\mathbf{for}$ loop and performs certain mutations on the 
highest-priority test case $\mathbf{c}$ taken from $\mathcal{Q}$, shown at lines 1-2. 
The mutation forms a set of new test cases $\mathcal{M}_\mathbf{c}$. 
Next, at lines 3-4, our algorithm starts the simulation by calling function 
$SimRun(\mathbf{c}')$ on each element $\mathbf{c}'$ in $\mathcal{M}_\mathbf{c}$. It 
also retrieves the safety issue report $isCrash(\mathbf{c}')$, the trajectory coverage 
result $drivingTraj(\mathbf{c}')$, and the liability decision $atFault(\mathbf{c}')$, 
respectively.  
If the $drivingTraj(\mathbf{c}')$ reveals that the autonomous driving vehicle reaches 
new road locations compared to historical data, our algorithm treats $\mathbf{c}'$ as 
valuable candidate and puts it into the priority queue $\mathcal{Q}$, shown at lines 5-6. 
Otherwise, at lines 7-8, our algorithm examines if the autonomous driving vehicle 
encounters a safety issue and is not at-fault by checking the results of 
$isCrash(\mathbf{c}')$ and $atFault(\mathbf{c}')$. If both conditions satisfy, 
our fuzzing technique detects a potentially risky case and stores $\mathbf{c}'$ for 
investigation.

Note that Algorithm~\ref{algo:search} compliments Fig.~\ref{fig:framework} with the 
formal descriptions. However, for the brevity reason, we omit the details of the 
functions in the algorithm. We leave the technical explanations of these functions
in Section~\ref{sec:asf-a}.


\begin{algorithm}\caption{Trajectory coverage-driven fuzzing} \label{algo:search}
\SetKwInput{Initialization}{Initialization}
\Initialization{\\
 \quad Priority queue $\mathcal{Q}$: initial random test cases;}
 \For{\textup{highest priority case} $\mathbf{c}\in\mathcal{Q}$}{
$\mathcal{M}_\mathbf{c} = mutate(\mathbf{c})$;\\
\For{\textup{mutated case} $\mathbf{c}'\in\mathcal{M}_\mathbf{c}$}{
$isCrash(\mathbf{c}'), drivingTraj(\mathbf{c}'), atFault(\mathbf{c}') = SimRun(\mathbf{c}')$;\\
\If{$drivingTraj(\mathbf{c}')$ contains new locations}{
  $\mathcal{Q}.insert(\mathbf{c}')$;}
 \If{$isCrash(\mathbf{c}') \& atFault(\mathbf{c}')$}{
 Store risky case $\mathbf{c}'$;}
}
}
 
\end{algorithm}

\subsection{Core techniques in the fuzzing engine}
\label{sec:asf-a}

\para{Trajectory coverage-based feedback mechanism.}
As analyzed in Section~\ref{sec:fuzzing}, the primary challenge of virtual safety fuzzing 
stems from the feedback mechanism and the measuring metric that drive the fuzzing progress. 
However, quantifying safety issues remains an open question and there is no standard way 
to make a general quantification. Alternatively, we propose a novel trajectory coverage as 
the structural metric for virtual safety fuzzing. The trajectory refers to the waypoints
that the autonomous driving vehicle has driven on the road in a simulation scene. 
The intuition is that by covering more areas on the map, we hope the self-driving  car  could  reach  some  corner  cases,  so  as  to increase the probability of triggering safety issues eventually.
Therefore, if \sysname~receives the feedback that the simulated vehicle in the virtual scenario drives a new road trajectory under a specific test case, \sysname~will give priority to this case for further mutations. 


\para{Crash signal processing}
Program execution crash, e.g., segmentation fault, presents a explicit crash \emph{signal} 
to the software fuzzer (ref. Fig.~\ref{fig:flow}). 
Analogously, we can consider all collisions as the "crash" signals for virtual safety
fuzzing. However, accident reports showed that the autonomous driving vehicle may not always 
be at fault in the collisions or accidents it involved. 
Reckless driving from nearby vehicles can cause unavoidable accidents despite the autonomous 
driving system behaves correctly. Thus those collisions should not be categorized into 
safety violations from the autonomous driving system.
\sysname~tackles this problem by a liability judgement against captured collisions. 
Specifically, \sysname~invokes a callback mechanism to decide the party at fault once 
agents like vehicles and stationary objects controlled by the simulator collide. 
For example, {\em Listing}~\ref{lst:crashsignal} shows a callback function in SVL simulator
on a collision. The function arguments include involved parties and collision location. Once 
\sysname~monitors the signal in simulation, it analyzes both the callback data and navigation 
history data before the collision to compute the at-fault parties, and makes decisions accordingly.

\begin{lstlisting}[language=Python, caption={SVL collision callback function}, label={lst:crashsignal}, basicstyle=\small ] 
def on_collision(agent1, agent2, contact):
  ...  
  print("{} collided with {} at {}"
    .format(name1, name2, contact))
ego.on_collision(on_collision)
\end{lstlisting} 

\para{Mutation strategies.} 
The mutation strategies in \sysname~focus on mutating the content of dynamic configurations
of a test case. Intuitively, fuzzing the JSON test case is close to the existing software 
fuzzing works on well formatted files like PDF, XML, YAML, etc. However, these existing 
works primarily concentrate on finding the implementation bugs in file interpreter or parser. 
By contrast, \sysname~needs to provide diverse yet valid test case files to set up simulation 
environments, so as to test behavioral differences of autonomous driving vehicles.
Note that each dynamic configuration element may correspond to a specific object in simulation. 
Fuzzing the element content has to take its semantics into consideration to produce meaningful 
mutants. Thus, we cannot simply adopt the AFL~\cite{AFL} strategies or the structure-aware 
fuzzing strategies~\cite{BlazytkoASASWH19,YouLMPZ019}.
To accommodate the the virtual safety fuzzing nature, we design a set of mutation strategies
in \sysname~as follows:
 


\begin{itemize}
    \item Arithmetic. To derive more trajectories on the map in a deterministic way, we attempt to 
    subtly increment or decrement numerical parameters in the configuration. An example is to add 
    or subtract $x$ coordinate and $y$ coordinate values of an agent's position.
    
    \item Flip. Flip strategy alters the state of a controllable object (e.g., traffic light color). Flip action also applies to numerical parameters. For example, we design position flip that reflects the position of an agent with respect to the center point of the driving area.
    
    \item Random. Random strategy randomly replaces the numerical parameters or states with the 
    values within a certain working range.
    
    \item Insert. Insert strategy specifically works for the set of observable agents. It adds 
    specific agents along with the routing path.
\end{itemize}

\begin{figure}[h]
    \centering
    \begin{minipage}{0.49\linewidth}
    \centering
    \includegraphics[width=1\linewidth]{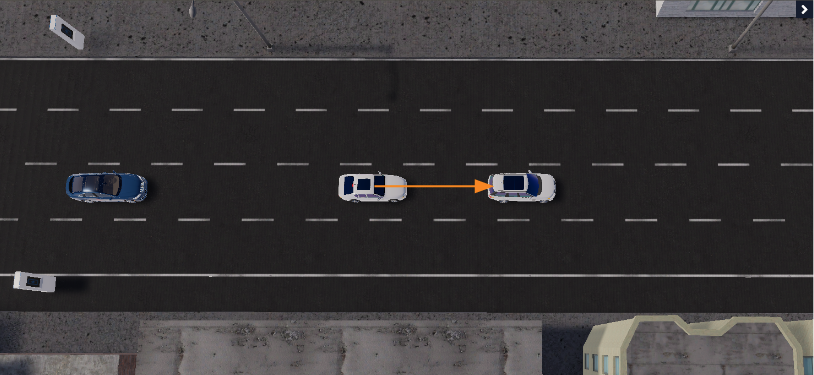}
    {Arithmetic}
    \end{minipage}%
    \begin{minipage}{0.491\linewidth}
    \centering
    \includegraphics[width=1\linewidth]{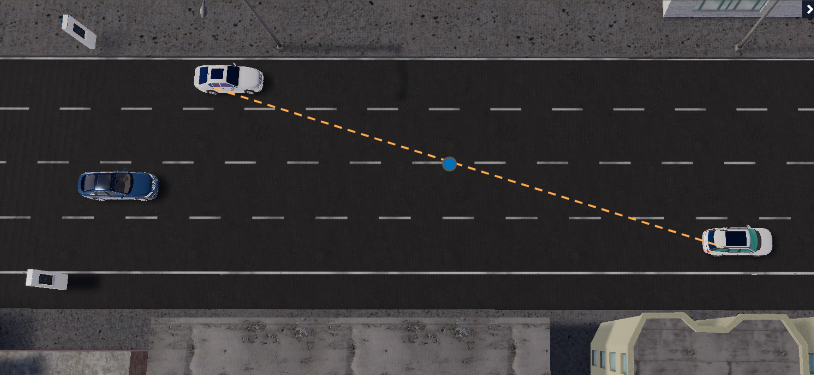}
    {Flip}
    \end{minipage}
    \centering
    \begin{minipage}{0.49\linewidth}
    \centering
    \includegraphics[width=1\linewidth]{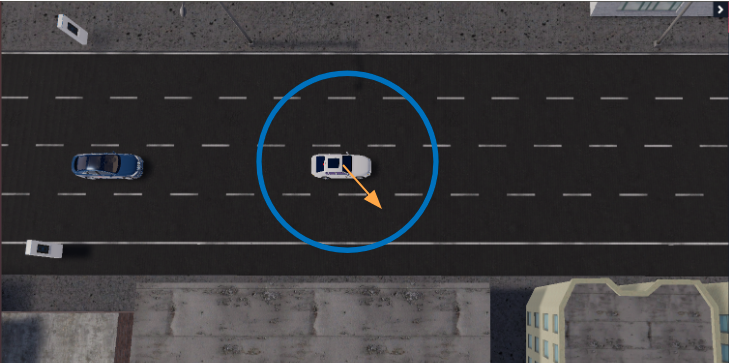}
    {Random}
    \end{minipage}%
    \begin{minipage}{0.49\linewidth}
    \centering
    \includegraphics[width=1\linewidth]{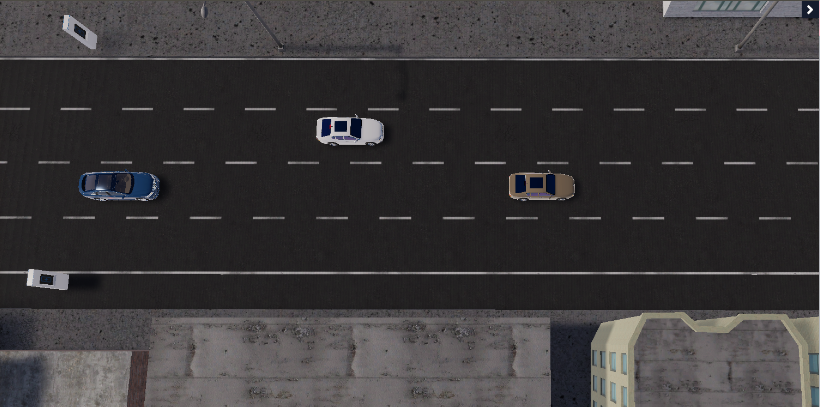}
    {Insert}
    \end{minipage}
    \caption{Examples of mutation strategies}\label{fig:mutation}
\end{figure}

Fig.~\ref{fig:mutation} demonstrates the visual effects of the above mutation strategies.
More workable mutation strategies will be tested and added to our strategy repository.

\section{Case Study}

To show the benefit of applying fuzzing on virtual test case generation, we use a concrete case study to demonstrate how \sysname~works. Through this case study, we will show how the coverage-driven feedback mechanism in \sysname~can help find potential risky test cases. We first demonstrate the trajectory coverage efficiency of \sysname, then present some risky cases found by \sysname. 
%

\subsection{Case study setup}

We make two definitions to depict the trajectory coverage.
\begin{definition}
\label{def1}
\textbf{Driving area}. The driving area on the map is a rectangle bounded by the specified 
geographic Universal Transverse Mercator (UTM) coordinates~\cite{langley1998utm} in the 
static configurations of a test case. 
\end{definition}

\begin{definition}
\label{def2}
\textbf{Trajectory coverage}. We divide the driving area into $1m\times1m$ blocks. The 
trajectory coverage presents the set of blocks covered by the autonomous driving 
vehicle trajectory in a simulation.
\end{definition}

In this study, we set the block granularity of the driving area as $1m\times1m$ for simplicity. 
Note that the vehicle may never reach certain areas within the driving area and the choice of 
block granularity might affect the search cost and experimental result. 
We leave the granularity reasoning as future work.

\begin{figure}[h]
    \centering
    \includegraphics[width=1\linewidth]{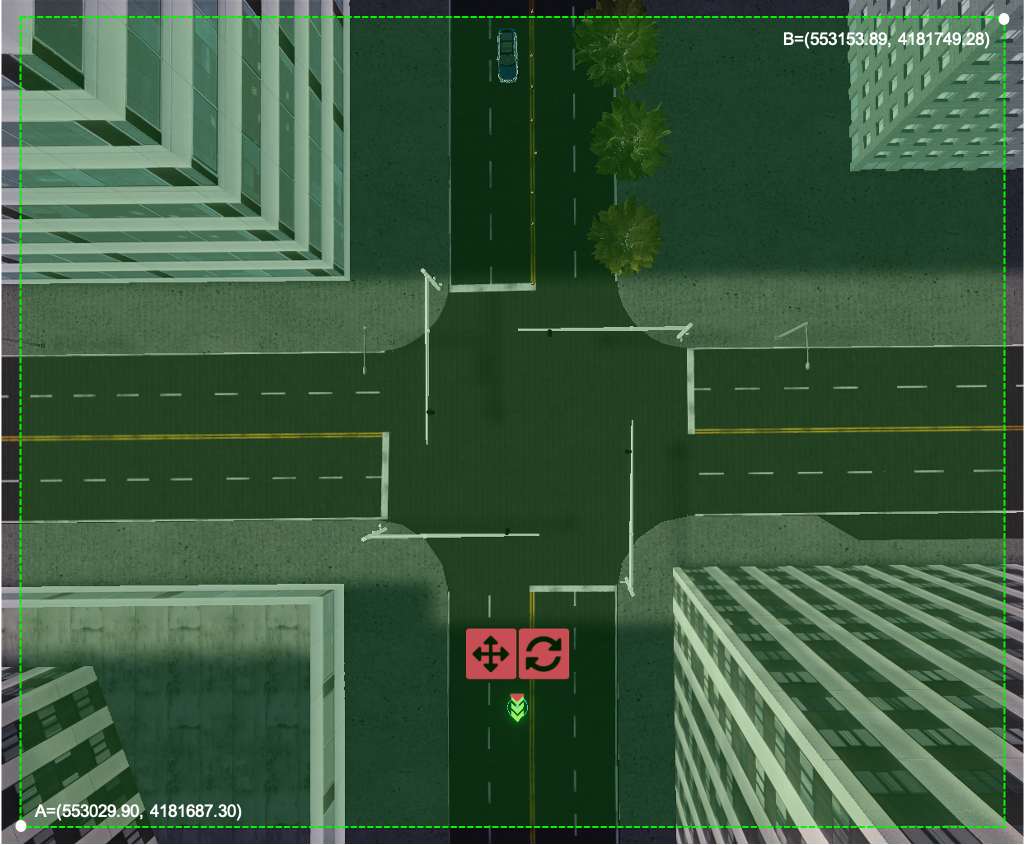}
    \caption{The driving scene in our case study}
    \label{fig:case}
\end{figure}

Fig.~\ref{fig:case} shows the basic driving scene in our study. 
The autonomous driving vehicle goes from the start point at the leftmost lane to the end point 
marked by the green arrows. The driving scene skeleton is formalized by the static configurations 
shown in Fig.~\ref{fig:JSON}-(a). The driving area in this case is the the green translucent
rectangle in Fig.~\ref{fig:case}), bounded by point  
$A=(easting=553029.90,northing=4181687.30)$ and point $B=(easting=553153.89,northing=4181749.28)$ 
in the UTM coordinate system. 
In the middle there is also an intersection with traffic lights.
We conduct all the experiments on a Ubuntu Desktop with an AMD Ryzen 7 
3700X CPU, 16GB memory, and an NVIDIA GTX1080 TI.

\subsection{Coverage efficiency}

\begin{figure}[h]
    \begin{center}
    \includegraphics[width=1\linewidth]{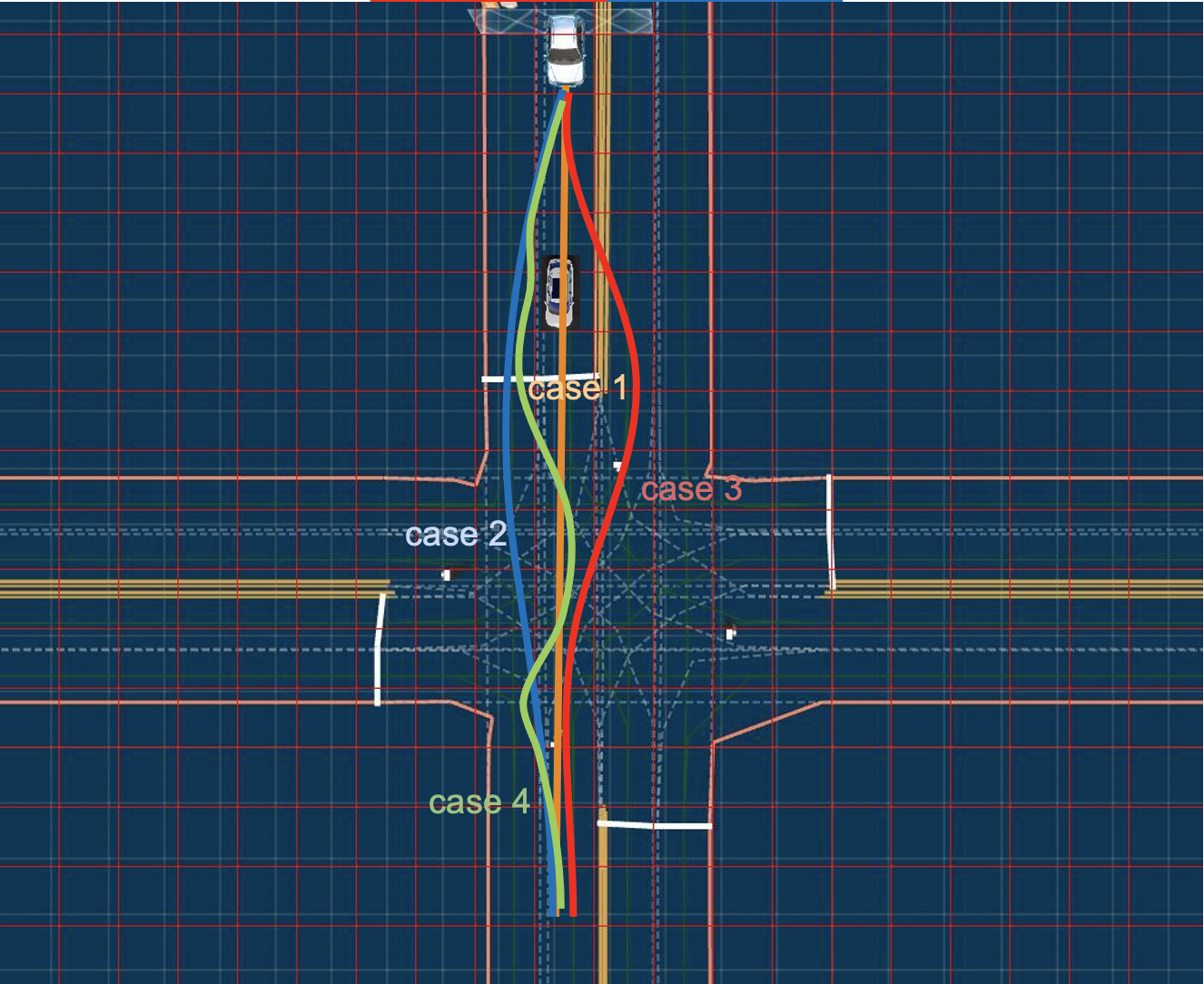}
    \caption{Trajectory coverage}\label{fig:coverage}
    \end{center}
\end{figure}

We use Fig.~\ref{fig:coverage} to visualize the concept of trajectory coverage. It shows 
the planar graph of the driving area, divided by the red $1m\times1m$ blocks. There are 
four driving trajectories generated by running an initial test case (case 1) and three 
mutated test cases (cases 2-4). We observe that case 2 and case 3 contribute to new 
trajectory coverage since both of them have covered new map blocks. By comparison, case 
4 covers no new blocks compared to other three, hence it would be discarded without 
further mutation in \sysname.

In this study, the mutating content is the positions of the surrounding NPC agents (e.g., the brown sedan in 
Fig.~\ref{fig:coverage}).
%
Now, we compare \sysname~with two other fuzzers regarding the ability of covering more 
driving area blocks. The first one is a random fuzzer that generates random positions 
of the NPC agents within the driving area. The second fuzzer is the state-of-the-art 
autonomous vehicle safety fuzzing tool AV-Fuzzer~\cite{9251068}. AV-Fuzzer developed a 
genetic algorithm driven by a fitness function, i.e., the maximum distance the vehicle 
can move without colliding with any obstacle or vehicles.

\begin{figure}[h]
    \centering
    \includegraphics[width=1\linewidth]{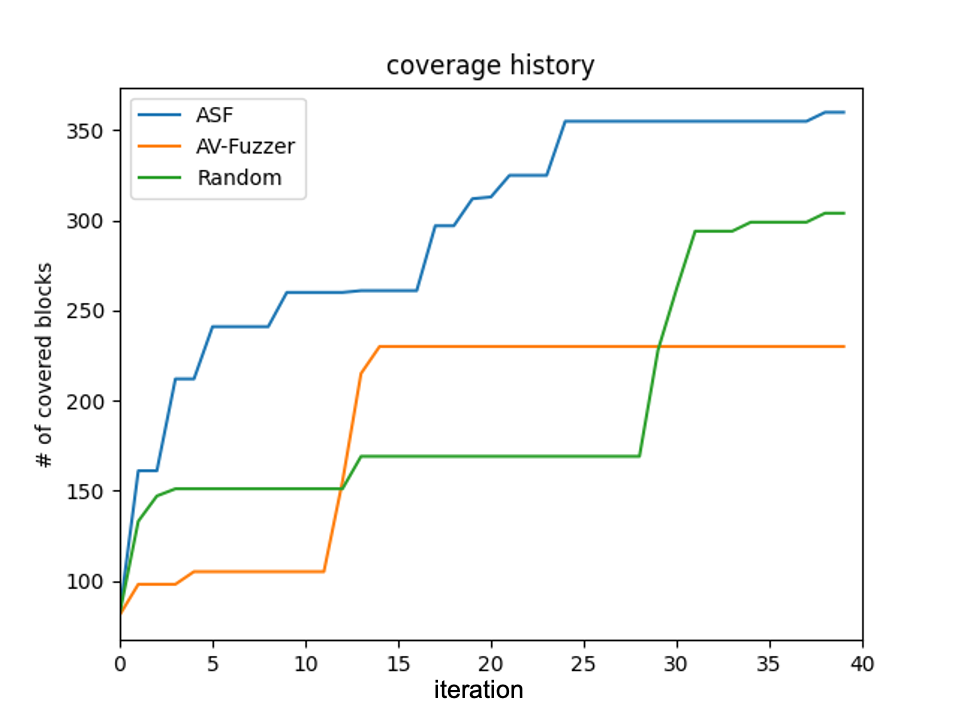}
    \caption{Coverage efficiency comparison}
    \label{fig:mullane}
\end{figure}

Fig.~\ref{fig:mullane} compares the aggregate number of covered blocks by \sysname, 
random fuzzer, and AV-Fuzzer in a 40-iteration simulation. The overall comparison 
shows that \sysname~covers obviously more drive area blocks than the other two 
approaches, resulting in a higher trajectory coverage. 
Moreover, in \sysname, the coverage growth rate at the early stage is significantly 
ahead of the other two methods, which indicates the superior fuzzing efficiency in 
\sysname~with the testing budget.

The reason that \sysname~outperforms both AV-Fuzzer and random fuzzer is two-sided. First, 
\sysname's feedback mechanism favors seeds that always contribute new trajectories. Second, 
\sysname~also conducts a preferential search near the seeds through the dedicated mutation 
strategies for virtual safety fuzzing.

\subsection{Risky cases}

%
Table~\ref{tab:scene} lists the statistics of the risky cases found by random fuzzer, 
AV-Fuzzer, and \sysname, in the 40-iteration simulation. The entries highlighted by red 
color are severe cases that lead to collisions. The orange-colored entries represent the 
near-collision cases.
As seen, within the 40-iteration simulation, \sysname~found one near-collision case and 
two risky cases that directly lead to collisions that hit a traffic cone and a sedan 
vehicle.
In comparison, AV-Fuzzer only found one near-collision case and the random fuzzer found 
no risky case.


\newcolumntype{o}{>{\columncolor{YellowOrange}}c}
\begin{table}[h]
    \caption{Risky cases found in the 40-iteration  simulation}
    \centering
    \begin{tabular}{|c|c|o|o|o|} \hline
        \rowcolor{Gray}
        \textbf{method} &\textbf{\# of risky cases}
        &\textbf{distance} &\textbf{hit speed} &\textbf{hit object}\\ \hline
        \rowcolor{White}
        RANDOM &0 &8.92 m &0 km/h & n/a \\ \hline
        AV-Fuzzer & 1 near-collision &3.57 m &0 km/h  & n/a\\ \hline
        \multirow{3}{*}{\sysname} &\multirow{3}{7em}{3 (2 collisions + 1 near-collision)}  &\cellcolor{OrangeRed} 0 m &\cellcolor{OrangeRed} 11.27 km/h &\cellcolor{OrangeRed} TrafficCone\\\cline{3-5}
         & &\cellcolor{OrangeRed} 0 m &\cellcolor{OrangeRed} 8.17 km/h &\cellcolor{OrangeRed} sedan\\ \cline{3-5}
         & &1.92 m &0 km/h & n/a\\ \hline
    \end{tabular}
    \label{tab:scene}
\end{table}

\begin{figure}[h]
    \centering
    \begin{minipage}{0.49\linewidth}
    \centering
    \includegraphics[height=0.8\linewidth]{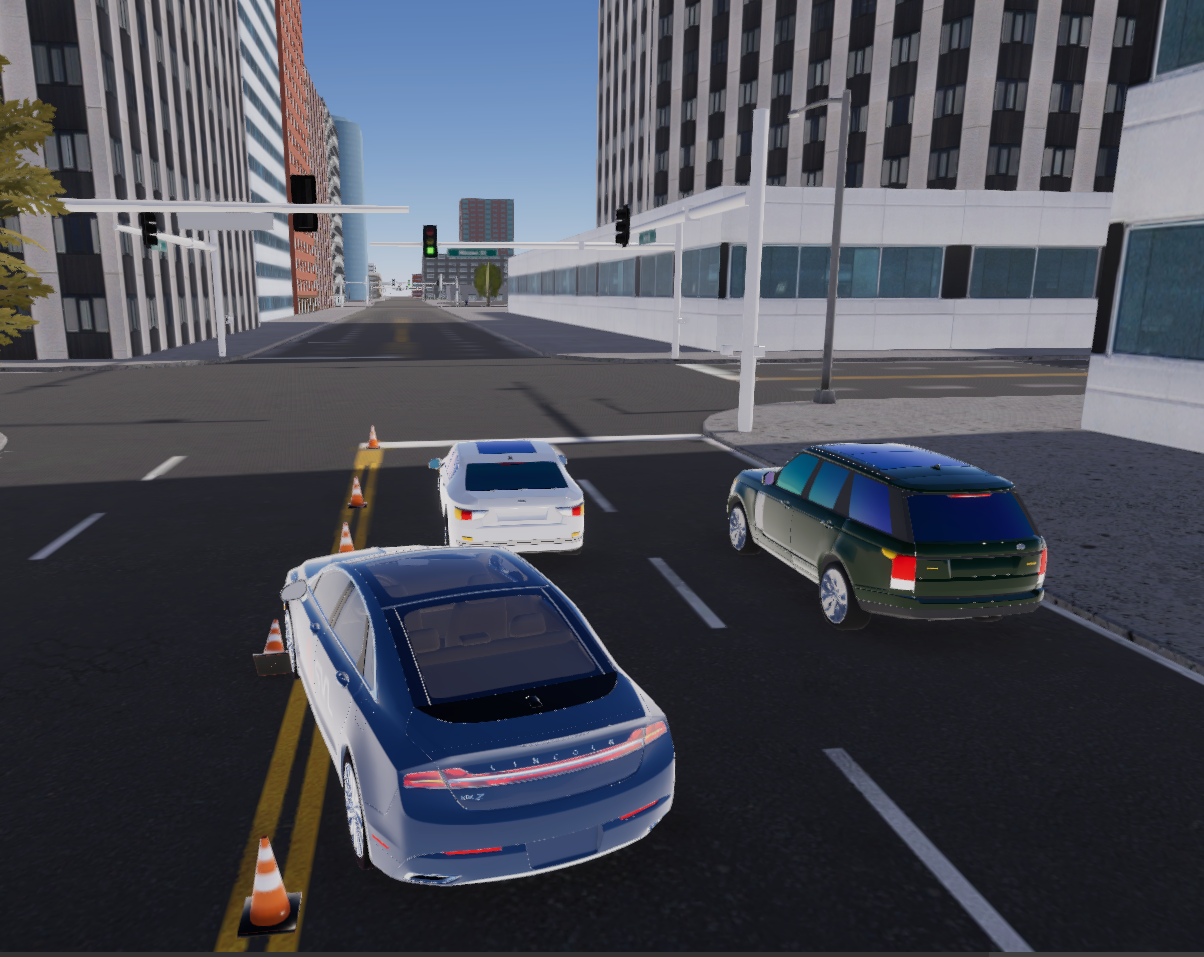}
    {simulation}
    \end{minipage}%
    \begin{minipage}{0.49\linewidth}
    \centering
    \includegraphics[height=0.8\linewidth]{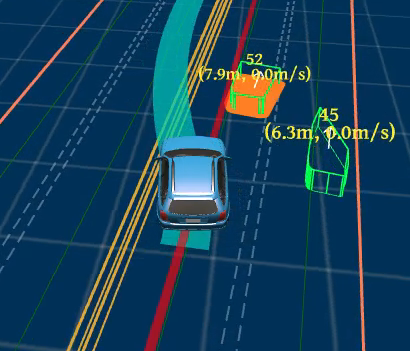}
    {planned trajectory}
    \end{minipage}
    \caption{Collision with traffic cone}
    \label{fig:cone}
\end{figure}

For better understanding of the collision found by \sysname, we visualize the traffic 
cone accident in Fig.~\ref{fig:cone}.
In the left-side simulation snapshot, the autonomous driving vehicle hit a traffic cone 
on the solid double-yellow line. 
In the right-side Dreamview UI snapshot, we can see that the autonomous driving car 
controlled by Apollo planned a trajectory that goes across the double-yellow line 
and points to the reverse lane on the left.
After analyzing the Apollo source code, we believe that the root cause of the planning
error works as follows. 
When there are surrounding stopping obstacles, i.e., the front NPC vehicle and the right-side
NPC vehicle, Apollo makes a decision to borrow the lane on the left. However, at this time, 
Apollo actually does not check the boundary type of the lane to be borrowed. As a result, 
it plans a trajectory across the middle double-yellow line, which forms a severe traffic violation and causes a collision in this case. This example shows that \sysname~can help to find subtle logical  flaws in autonomous driving system  that  would be  hard  to  detect  if  only  using  random or handcrafted test cases.
%


\subsection{Discussion}
Experimental results have shown that the coverage-driven feedback mechanism in 
\sysname~identifies more risky cases under the same driving scenario. We propose 
speculative explanations for the results.
One point is that when the trajectory of an autonomous driving vehicle covers more 
driving area, there would be more possibilities to trigger dangerous corner cases, 
such as the above traffic cone accident.
The fitness-driven AV-Fuzzer searches for the \emph{high potential} seeds that might 
lead to collision. However, it can be trapped in the local optimum like the 
near-collision case. Though the restart mechanism can help AV-Fuzzer get out of the 
local optimum, its efficiency of finding new trajectory is close to that of a random 
fuzzer. 
On the other hand, coverage-driven \sysname~explores different trajectories as much 
as possible. Though these trajectories may not necessarily lead to collisions, they
help \sysname~to filter the benign cases thus approaching the severe cases. 



\section{Conclusion and Future Work}

Just as software testing cannot primarily rely on human efforts to find bugs, virtual safety 
testing also desires automated test case generation due to the huge configuration 
search space. 
Toward this end, we propose a coverage-driven fuzzing method \sysname~to generate high-quality 
test cases for the virtual safety testing of autonomous driving system.
\sysname~can automatically mutate dynamic configurations to form new test cases for detecting risky 
autonomous driving scenes in simulation. 
Comparisons among \sysname, the state-of-the-art AV-Fuzzer, and the random fuzzer show 
that \sysname~can find more severe safety issues.
With \sysname, we have found few flaws in a popular open-source autonomous driving system in a short time. 

To systematically study virtual safety testing, we plan to explore the following points in the future: 
\begin{itemize}
    \item How to improve the feedback mechanism to make the autonomous driving car be more prone 
    to collisions?
    
    \item How to design appropriate environmental settings for virtual safety fuzzing, such as 
    driving area and block granularity, in order to balance the search space and the fuzzing cost?
    
    \item How to optimize the combination of various mutation strategies to improve the fuzzing efficiency?
\end{itemize}

We are actively conducting research to answer these questions and pursuing approaches to generate 
more efficient virtual drive test solutions. 

\bibliographystyle{plain}
\bibliography{ref}
%



\end{document}